\documentclass{article}

\usepackage{arxiv}

\usepackage[utf8]{inputenc} 
\usepackage[T1]{fontenc}    
\usepackage{hyperref}       
\usepackage{url}            
\usepackage{booktabs}       
\usepackage{amsfonts}       
\usepackage{nicefrac}      
\usepackage{comment} 
\usepackage{microtype}      
\usepackage{lipsum}
\usepackage{graphicx}
\usepackage[round]{natbib}

\usepackage{atbegshi}
\AtBeginDocument{\AtBeginShipoutNext{\AtBeginShipoutDiscard}}
\title{From Idea to Implementation: Evaluating the Influence of Large Language Models in Software Development - An Opinion Paper}

\author{
  Sargam Yadav \\
School of Informatics and Creative Arts\\ 
Dundalk Institute of Technology,\\ Dundalk, Ireland \\
  \texttt{d00263026@student.dkit.ie}  \\ 
  \And
Asifa Mehmood Qureshi \\
School of Informatics and Creative Arts\\ Dundalk Institute of Technology,\\ Dundalk, Ireland\\
  \texttt{d00273262@student.dkit.ie} \\ 
   \And
Abhishek Kaushik\thanks{corresponding author} \\
School of Informatics and Creative Arts \\ Dundalk Institute of Technology,\\ Dundalk, Ireland\\
  \texttt{abhishek.kaushik@dkit.ie} \\
  \And
Shubham Sharma \\
The Centre for Research in Engineering Surface Technology (CREST)\\ TU Dublin\\ Dublin, Ireland\\
  \texttt{Shubham.Sharma@TUDublin.ie} \\
   \And
Roisin Loughran \\
School of Informatics and Creative Arts\\ Dundalk Institute of Technology,\\ Dundalk, Ireland\\
  \texttt{roisin.loughran@dkit.ie} \\
 \And
Subramaniam Kazhuparambil \\
Zendesk,\\ Dublin, Ireland\\
  \texttt{subu.kazhuparambil@zendesk.com} \\
    \And
Andrew Shaw \\
Department of Computing Science and Mathematics\\ Dundalk Institute of Technology,\\Dundalk, Ireland\\
  \texttt{andrew.shah@dkit.ie} \\
  \And
Mohammed Sabry \\
ADAPT Centre\\\ Dublin, Ireland\\
  \texttt{mohammed.sabry@adaptcentre.ie} \\
  \And
Niamh St John Lynch \\
School of Informatics and Creative Arts\\ Dundalk Institute of Technology,\\ Dundalk, Ireland\\
  \texttt{niamh.stjohnlynch@dkit.ie} \\
  \And
Nikhil Singh \\
National College of Ireland,\\ Dublin, Ireland \\
  \texttt{nikhil\_singh@outlook.com} \\
  \And
Padraic O'Hara \\
School of Informatics and Creative Arts\\ Dundalk Institute of Technology,\\ Dundalk, Ireland\\
\texttt{d00273238@student.dkit.ie} \\
 \And
Pranay Jaiswal \\
School of Informatics and Creative Arts\\ Dundalk Institute of Technology,\\ Dundalk, Ireland\\
\texttt{d00273754@student.dkit.ie}
  \And
Roshan Chandru \\
School of Informatics and Creative Arts\\ Dundalk Institute of Technology,\\ Dundalk, Ireland\\
\texttt{roshan36691@gmail.com} \\
  \And
David Lillis \\
School of Computer Science,\\ University College Dublin (UCD), \\ Dublin, Ireland\\
\texttt{david.lillis@ucd.ie} \\
}

\begin{document}
\maketitle

\begin{abstract}
The introduction of transformer architecture was a turning point in Natural Language Processing (NLP). Models based on the transformer architecture such as Bidirectional Encoder Representations from Transformers (BERT) and Generative Pre-Trained Transformer (GPT) have gained widespread popularity in various applications such as software development and education. The availability of Large Language Models (LLMs) such as ChatGPT and Bard to the general public has showcased the tremendous potential of these models and encouraged their integration into various domains such as software development for tasks such as code generation, debugging, and documentation generation. In this study, opinions from 11 experts regarding their experience with LLMs for software development have been gathered and analysed to draw insights that can guide successful and responsible integration. The overall opinion of the experts is positive, with the experts identifying advantages such as increase in productivity and reduced coding time. Potential concerns and challenges such as risk of over-dependence and ethical considerations have also been highlighted. 
\end{abstract}


%


\maketitle

 \section{Introduction}

The swift evolution of artificial intelligence in the past decade has allowed advanced Natural Language Processing (NLP) tools to perform tasks such as translation, summarization, and information retrieval. These tools are becoming more easily available and integrate seamlessly with our daily lives in the form of virtual assistants. Large Language Models (LLMs) are large and complex deep learning models with millions or billions of parameters trained on vast amounts of data. Models such as ChatGPT can carry out human-like conversations keeping context in mind, highlighting their advanced Natural Language Understanding (NLU) \citep{brown2020language} and Natural Language Generation (NLG) \citep{dong2022survey} capabilities. Transformer-based models such as the Bidirectional Encoder Representations from Transformers (BERT) \citep{devlin2018bert} and Generative Pre-Trained Transformer (GPT) \citep{radford2018improving} have demonstrated unparalleled capabilities in tasks such as text generation \citep{luo2022biogpt}, named entity recognition \citep{liu2021ner}, machine translation \citep{clinchant2019use}, text classification \citep{yadav2023comparative, kaur2019cooking, yadav2022disruptive, shah2020opinion}, and more. LLMs are pre-trained on large amounts of data and can gain considerable amount of domain knowledge, which allows them to be a valuable assistant in fields such as software development \citep{hornemalm2023chatgpt} healthcare \citep{biswas2023role} education \citep{adeshola2023opportunities, kaushik2025exploring}, and agriculture \citep{10682965, 10306134,10.1007/978-981-99-0981-0_21, kovvuri2023disruptive}. Since their launch, LLMs such as GPT-4 \citep{achiam2023gpt}, Codex \citep{chen2021evaluating},and CodeBERT \citep{feng2020codebert}, have been utilised by researchers and practitioners to efficiently perform various tasks in the software development workflow such as code completion \citep{raychev2014code}, code synthesis from natural language \citep{desai2016program}, debugging, code review \citep{zhao2023right}, code comment and documentation generation \citep{pu2023summarization}, among others. 

In the context of software development, LLMs can greatly reduce workload for developers, allowing them to focus on more creative aspects of the coding process \citep{githubcopilot}. They can streamline the development process by ensuring compliance and version control, improving code quality, and promoting collaboration \citep{zhang2024llm}. They can also be leveraged to develop interactive educational tools that can provide tailored feedback to programming students \cite{lyu2024evaluating} \cite{liffiton2023codehelp}. Table \ref{tab:lit_review} displays a list of the commonly used LLMs for software development, along with their parameters, training data, training language, and architecture.

\begin{table}[!h]
    \centering
    \small
    \begin{tabular}{ccccc}
    \hline
       Model  & Parameters & Training data & Language & Architecture  \\
         \hline
        CodeParrot & 1.5b & GitHub  & Python & GPT-2  \\
        GPT-Neo & 2.7b &Pile dataset \citep{gao2020pile} & -- &GPT-2 based \\
         GPT-NeoX & 20b &Pile dataset & --&GPT-3  \\
        GPT-J & 6b &Pile dataset & -- &GPT-3 \\
        Codex & 12b & GitHub, + & Python, + &GPT-3  \\
        CodeBERT & 125M& GitHub & Multiple &BERT  \\
        CuBERT & -- & GitHub & Python &BERT \\
        PolyCoder & 2.7 b & GitHub & 12 languages & GPT-2 \\
        
        \hline
        
    \end{tabular}
    \caption{Large Language Models for Software Development}
    \label{tab:lit_review}
\end{table}




However, there have also been certain risks and limitations associated with using LLMs. LLMs may not fully understand the context of a prompt \citep{deng2023pentestgpt}, leading to erroneous responses. Over-reliance on LLMs to perform tasks may inhibit problem solving and critical thinking skills. There are also several ethical considerations of using LLMs for software development such as unintended biases \cite{huang2023bias}, security risks \cite{yao2024survey}, transparency, and explainability \citep{liyanage2023ethical}. LLMs are trained on vast amounts of data which contains human-like biases \citep{tjuatja2023llms}. The models can reflect these biases in their responses and cause further harm. It is important to mitigate biases and address the concerns to ensure fairness in the model and its related outputs.



In this study, a group of software development experts, consisting of researchers and practitioners, was surveyed to understand the potential applications of LLMs in software development. Analysis of opinions obtained through a loosely structured written interview from 11 experts, consisting of researchers and practitioners, has also been performed to draw insights about the potential of LLMs. The rest of the article is structured as follows: In Section \ref{motivation}, we discuss the motivation behind conducting this survey, as well as define the hypothesis and research questions. Section \ref{method} details the methodology used in the study, Section \ref{opinions} describes the opinions of experts regarding the integration of LLMs in processes such as code generation, review, debugging, and so on. Section \ref{overview} provides an overview and analysis of the expert opinions, as well as a discussion of the research questions. Section \ref{conclusion} concludes the review study.

\section{Motivation} \label{motivation}
LLMs such as GPT, CodeBERT, and Codex have demonstrated transformative potential in software development. The hype surrounding LLMs creates ambiguity regarding their potential applications and drawbacks. There is some concern about job security, as it is perceived by some that models will become better at performing software development tasks than professional developers \cite{kuhail2024will}. Although LLMs can be an invaluable tool for learning, students may also use it unfairly and compromise academic integrity \cite{perkins2023academic} \cite{pudasaini2024survey}. Additionally, as with any new technology, there may be a steep learning curve before it can be seamlessly integrated \cite{kirova2024software}. In order to get a clear picture of the benefits and drawbacks of LLMs in the field of software development, it is important to analyse the opinions of experts working in the field. The motivation behind conducting this study is to analyse and perform thematic coding on written interviews obtained from 11 experts, including researchers and practitioners, in the area of machine learning and artificial intelligence about the use of LLMs in software development. 

The hypothesis of the survey is as follows:

\textit{Hypothesis: Experts such as researchers and practitioners have a positive outlook towards the integration of LLMs in software development.}

The research questions structured to explore the hypothesis are as follows:

\begin{enumerate}
    \item RQ1: What are the applications and advantages of LLMs in software development as per expert opinions?
    \item RQ2: What are the limitations and potential concerns of LLMs in software development as per expert opinions? 
\end{enumerate}

\section{Methodology} \label{method}

The study has been performed by obtaining loosely-structured written non-risk interviews from experts which include practitioners and researchers. Practitioners are individuals who are currently working with software development in industry and researchers are working on LLMs and other related areas. The search for the experts was done through Google Scholar, by contacting fellow researchers in the laboratory, and scheduling meetings with acquaintances. 36 experts, 18 males and 18 females, were contacted across the globe to get their opinions on the use of LLMs in software development. The study contains the opinions of the 11 responses that were received. The experts are not familiar with each other and therefore group bias is avoided. Consent has been obtained from the experts to summarize and share their opinions, and reduce overlapping content. The interviews were non-risk and did not require the disclosure of any personal information from the interviewer.



Table \ref{tab:expert_data} displays the gender and role information for the experts interviewed in the study. There are a total of 3 female and 8 male experts. There are 4 practitioners and 7 researchers. Out of the male experts, 2 are practitioners and 6 are researchers. Out of the female experts, 2 are practitioners and 1 is a researcher. Thematic coding of the opinions is performed by two annotators, and inter-annotator agreement is calculated using Cohen's kappa in Section~\ref{overview}.

\begin{table}[!h]
\centering
\caption{List of the Experts with their Roles and Gender}
\begin{tabular}{|l|l|l|l|}
\hline
\textbf{No.} & \textbf{Role} & \textbf{Gender}\\
\hline
Expert 1  & Practitioner & Male       \\ 
Expert 2  & Practitioner & Female \\ 
Expert 3  & Researcher   & Male       \\  
Expert 4  & Researcher   & Male   \\ 
Expert 5  & Researcher   & Female \\ 
Expert 6 & Researcher   & Male   \\ 
Expert 7 & Researcher   & Male        \\ 
Expert 8 & Researcher   &   Male     \\ 
Expert 9 & Practitioner & Female \\ 
Expert 10 &  Researcher            &  Male      \\ 
Expert 11 & Practitioner & Male \\ \hline
\end{tabular}
\label{tab:expert_data}
\end{table}

\section{Expert Opinions} \label{opinions}

In this section, the opinions of the experts have been analysed and categorized based on different aspects such as role of LLMs in code generation and review, natural language understanding, quality assurance, predictive analysis, developer collaboration, as well as bias and ethical consideration of LLMs for code generation. The experts were given these subheadings as a prompt, and some responded to one or more of them. The content from the unstructured essays was classified into the mentioned classes without any modifications and rephrasing.

\subsection{The Role of LLMs in Code Generation}
LLMs are extensively used in code generation activities. These language models have the capability to generate accurate and efficient code. This section discusses the opinions of the experts on utilizing LLMs in code generation tasks.

\textit{Expert 1:} The advent of LLMs has rapidly accelerated software development. Getting started with a new library, new programming language, or a new tech stack in general has never been easier. A smart use of LLMs employed by our team is translating code across different languages. As a team of Scala Engineers, we had to develop a new glue service in Python and for some of the nuances, we wrote Scala code and translated it into Python using LLMs. While there is ease of use, the obvious caveat when using such models is that the burden of proof is on the user. However, in the case of software development, the accuracy of such models can be easily and fairly validated as if the code generated is correct, there should not be any compilation errors. Moreover, unit, split, and integration tests should validate that the business logic is preserved.

\textit{Expert 4:} To illustrate the effectiveness of LLMs in code generation, a few case studies and examples are examined. GitHub Copilot \citep{githubcopilot} powered by OpenAI’s Codex, is an example of LLMs in Code generation. Using Copilot, developers can automatically generate code based on their comments in the natural language. It can even be corrected in real-time by entering and modifying the text, which significantly speeds up code writing and adds more efficiency. LLMs have also been used to translate code between different programming languages \cite{hindle2016naturalness}. Language and cross tools problems can be overcome using source code translation integration. For example, if the code snippet is written in English, it can be transformed into Python, Java, or any other language. This facilitates cross-language collaboration and code reuse, particularly in diverse development environments. However, LLMs' potential role in code generation is part of the picture, and some issues and obstacles must be addressed. Communication is an innate human skill, but when people talk to one another, sometimes the language they use is unclear, such that it has more than one interpretation. It is almost impossible to implement such instructions in a program. In some cases, large language models might not be able to grasp the user’s ambiguous instructions, causing them to generate \citep{chen2021evaluating} code that is not desired. Since LLMs take data from the wide range of texts on which they are trained, they may not always include highly specialized or domain-specific programming concepts. This problem hinders LLMs' ability to generate codes that follow specific standards or advanced programming.

\par \textit{Expert 8:}  The tasks like code generation, creating summaries, and solving problems
are easily handled by LLMs. They are mainly used in code development to
automatically generate complex code. This can save developers time and effort.
LLMs has ability to understand different programming languages and can produce accurate code snippets. This helps developers by redimproving productivity and enhancing code quality, which will help in reducing errors. By this, developers can focus on high-level design and problem-solving tasks with more efficientdevelopment processes \citep{ozkaya2023application}. LLMs such as Codex and GPT-4 show good abilities in summarizing, creating, understanding code and they can solve complex programming problems effectively. They efficiently address intricate programming issues with a notable success rate. LLMs can also help in speeding up the development process using the features
like code completion. LLMs can make coding more efficient by reorganizing
it and can work directly in development environments to provide real-time help
and support by analyzing code. In addition to this, LLMs can be customized for specific projects, which makes them versatile tools in software development.
The code generation is made much easier by automating and streamlining the
coding process through advanced models like Codex. Through evaluating these
model’s capability to produce correct and efficient code snippets from prompts
shows that these AI tools have the potential to change software development,
making coding tasks faster and easier for developers \citep{chen2021evaluating}. Also, \citep{koziolek2023llm}, present case studies showing the effective use of LLMs for producing control code from piping and instrumentation diagrams in an industrial setting. The study uses reference piping and instrumentation diagrams to test the LLMs' ability to identify controllers and generate control code. The LLMs were able to generate plausible interlock code and startup sequences despite challenges like misinterpretation of graphical elements and misaligned signal references. It is found that the generated code required adjustments for practical use, highlighting the need for more contextual information in the prompts to improve accuracy. The other case study focuses on control logic generation for a butane regeneration system. This study uses a piping and instrumentation diagrams. This method faced challenges with hallucinated controllers and inaccuracies in identifying equipment. But it is found that by focusing on a specific diagram section and refining prompts, it successfully generated syntactically correct structured text code for piping and instrumentation diagrams control loops and startup sequences. This case study illustrated the importance of detailed prompts and the potential for LLMs to produce functional code when given concrete operational parameters and corrected for identified errors. 

\textit{Expert 10:} Other emerging approaches to augmenting and improving LLMs may also find applications in code generation, such as Retrieval Augmented Generation (RAG) and Chain of Thought to improve code generation \citep{li2023structured}.  RAG allows more verifiable and predictive results by searching a data source to prime the prompt with additional data. Mixed models which route to a collection of more specialized models are being developed, for instance, Mixtral \citep{jiang2024mixtral}.  The result of many of these developments is that models are now getting better at the same time as getting smaller and faster to train. The recent release of Gemini 1.5 shows significant advancement in multi-mode models, integrating video, image as well as language. Tools are already coming to market that use AI to convert design directly to code, without the need for any conventional coding. A recent example of this is Builder.io, which takes web designs in Figma documents and converts them directly to code, producing HTML, React, and Javascript just as a developer would \citep{Builder}. Another unfolding space is the generation of No Code AI Application Development.  These tools allow a user to generate business solutions by integrating AI models \citep{liu2024empirical}. An example of this tool is Imagica, which provides a graphical interface that allows users to specify and configure AI models to meet business objectives without the need for any traditional code. There are other tools that allow the rapid integration of LLMs to build solutions, such as Langchain. Agents are another area of growth which have the potential to replace many tasks currently performed by code, or even replace the activity of developers themselves. The emergence of open source models has also significantly lowered the barriers to LLM model development, as demonstrated by Stanford with Alpaca model they trained from Llama at low cost \citep{taori2023alpaca}. This model also employed generative AI to produce training data. This opens up the possibility for companies to develop their specialized models which would produce a whole new area of software development and related tools. A further result of Open source models is the interest in running local LLMs. Not only can companies develop their own proprietary models, they can run on their own infrastructure. Beyond that, this looks set to power AI at the Edge as the next generation of processors should be able to run these models in mobile, desktop, and IoT devices. Along with the growth of tools for development, there is likely to be a significant evolution in AI tools for operations. This will include security tools, Infrastructure as Code, and improved integration and deployment tools. In addition, the tools necessary for training and deploying LLMs will require new solutions. AI Augmented Testing and Security tools have evolved rapidly alongside LLMs. These tools automate tasks that were previously manual, expensive, and required high levels of expertise. These tools also enhance the throughput of the Continuous Integration (CI)/Continuous Development (CD) pipeline, which will be essential as AI code generation is likely to increase the velocity and volume of new products. Indeed, every part of the software supply chain management should be enhanced to support productivity increases and the growth of new projects that were previously not commercially viable.

\subsection{Enhancing Natural Language Understanding for Developers}
LLMs use prompts that enhances communication and understanding of natural language thus providing leverage to developers. Expert opinions that discusses the application of LLMs in enhancing natural language understanding are discussed below:

\textit{Expert 1:} Companies like Amazon (Code Whisperer), Microsoft (through their collaboration with OpenAI and ChatGPT), Google, and Facebook (through their models Bard and Llama [Open Source] respectively) are active players in this space, there is a strong focus and investment towards enhancing developer experience and productivity. 

As LLMs get integrated with scrum software such as JIRA and Google Docs, they can be used for writing tasks such as documentation and planning sprints, without human errors, which can greatly benefit teams. Furthermore, such tools can greatly accelerate the onboarding of new engineers. For example, a new engineer can ask the LLM to explain a line of code or doc in the context of the entire org without having to ping individual engineers or get lost in the Confluence rabbit hole.

\textit{Expert 4:} Models like GPT-3 and BERT use advanced deep-learning approaches to comprehend and create text resembling human speech. This capability has been applied to transform casual user interaction in natural languages into working code snippets. \citep{brown2020language}. For example, a programmer can describe any function or algorithm in plain English, and the language model can easily change this description into a function or algorithm in any coding language required. The potential of bridging the gap between natural language and code helps developers to work with higher efficiency and productivity. Projects like GitHub Copilot demonstrate the potential of LLMs in real-world applications. Nevertheless, the questions that involve natural language ambiguity and the persistence of doubts about domain knowledge concerning applying the LLMs in code generation should be addressed.

\subsection{LLMs in Code Review and Quality Assurance}
This section highlights the application of LLM in code review and quality assurance in the opinion of experts.

\textit{Expert 1:} Add-ons such as Github co-pilot start being adopted at the enterprise level, and human-escaped bugs and Quality Assurance gaps can be filled with the model running additional checks to ensure business logic and data integrity/coherence. 

\textit{Expert 2:}  Code Reviews (CR) are essential in reducing risks arising from medical devices in development and once placed on the market.  The introduction of LLMs into the code review process is now an urgent necessity to ensure Quality Assurance is adequately supported and capable of identifying and removing potentially life-threatening risks to patients \citep{sametinger2015security}. Moreover, the potential benefits of introducing LLM for improvement in effectiveness and accuracy of the code review performed with impact on the early development lifecycle is significant, including: 
 a) best practice in coding standards readily available using automated tools; b) continual and automated consistency check of coding style and practices; c) automated commenting of code by the LLM, freeing up the coder to deal with complex tasks; d) automation of code review for vulnerability against known vulnerabilities and smells; e) removal of waste (human resources, time and cost) from manual reviews and improving the effectiveness of the review; f) enabling integration of static and dynamic code analysis with suggestions from the LLM for action by developers; g) promotion of learning and competency amongst coders, and h) avoidance of over-reliance on quality or validation functions to find the bugs, and i) lastly, self-learned adjustments from deep-learning with developer feedback mechanisms improve the performance of the LLM within an organisation, so that ‘skills’ are not lost to a competitor from resignation or retirement \citep{dam2016deep, madera2017case}. Also, the use of LLMs for CR is a necessity in a risk-adverse highly complex area of software development and with the ever-increasing software-related defects identified in medical devices, including recalls and adverse events. \citep{madera2017case} demonstrate the benefits of a prediction model for code reviews with high precision of 97\% and accuracy of 92\% for large-scale medical software projects considered worthy of implementation. The ability to demonstrate measurable improvements from the introduction of LLM in CR is self-evident in terms of project performance and most importantly, the potential for mitigation of clinical and patient risk.

\textit{Expert 8:}  LLMs help developers understand code by summarizing it, finding and fixing bugs, correcting errors, and LLMs can also help in speeding up the development process using the features like code completion. LLMs can make coding more efficient by reorganizing it and can work directly in development environments to provide real-time help and support by analyzing code.

\textit{Expert 11:} Code review (CR) is a vital part of software development. CR allows reviewers to identify defects and provide feedback, allowing developers to address the issues. Typically, the CR process consists of two steps: defect identification and repair. \cite{zhao2023right} highlight various LLM models that can be used for code reviews, namely  ChatGPT, InCoder, CodeT5+, CodeFuse, LLaMA, CodeGen-2, CodeLLaMA and CodeReviewer. All of these models can understand both natural language and programming languages, and they can also produce superior code snippets. Once the model is finalized, the second step is identifying the prompts. Prompts are input strings sent to a language model to specify the desired response \cite{lou2023prompt}. Prompts are the primary means by which users interact with LLMs. Prompt design has a significant impact on inferences drawn, so it is important to consider when using LLMs for specific tasks. It is critical to identify the optimal prompt that maximizes performance while reducing implementation effort. According to \cite{zhao2023right}, the following prompts can be used for these models.

\textit{“Fix the following buggy code snippet…”}

\textit{“Fix the following buggy code snippet according to the suggestion in the"//<Comment>" line.”}

As per \cite{Medium}, LLM offers the following advantages in the code review process.
\begin{enumerate}
    \item \textbf{Efficiency:} LLMs can review code faster and more accurately than human reviewers. This is especially important in large codebases, where manually reviewing every line of code can take a long time.

\item \textbf{Consistency:} Unlike humans, LLMs do not tire or lose concentration. They can consistently provide high-quality reviews, regardless of code volume or complexity.
\item \textbf{Continuous Learning:} As LLMs review additional code, they learn and grow. 

\end{enumerate}

This continuous learning process helps them improve their code review skills over time.

\subsection{Collaborative Software Development with LLMs}
The utilization of LLMs provides a platform for sharing and discussing ideas, reviewing code, and fostering a collaborative development environment. This section focuses on the collaborative aspects of LLMs from experts' perspectives.

\textit{Expert 1:} At scale, CI/CD workflows can greatly benefit from LLMs as they can create efficient testing methods via automatically generating mock data and associated Unit, split, and Integration tests just by feeding business logic and code to the model. Further, LLMs can set version control system standards across a company by enforcing descriptive commit messages and commit history. This can ease collaboration in large-scale organizations where many devs are working on the same code base parallelly. Project management and estimation can be vastly improved by feeding Objectives and Key Results (OKRs), financial targets, engineering and product capacity, and market and competition trends.

\textit{Expert 6:} One of the impactful techniques in software development is pair programming, which is where two programmers work together at one workstation. One programmer writes code, while the other reviews each line of code as it's typed in, providing feedback and thinking ahead strategically. Pair programming improves code quality, reduces development cycle time, and accelerates the feedback loop. The recent advancements in large neural networks, particularly those built on transformer-based architectures, have revolutionised code generation. These advancements have enabled the integration of such models into pair programming workflows. GitHub Copilot stands out as one of the most matured models in this domain, facilitating collaborative coding with its sophisticated capabilities. Also, in a large-scale survey conducted by the GitHub copilot team, they examined the benefits experienced by developers using GitHub Copilot. Key findings include:
\begin{itemize}
    \item Fast coding: Developers who used GitHub Copilot \citep{githubcopilot} completed the task significantly
faster–55\% faster than the developers who didn’t use GitHub Copilot.
\item Enhanced developer satisfaction: 60–75\% of users reported increased job fulfilment, reduced frustration while coding, and the ability to focus on more fulfilling tasks.
\item Conservation of mental energy: Users noted that GitHub Copilot helped maintain workflow (73\%) and saved mental effort during repetitive tasks (87\%), contributing to overall developer happiness. This is crucial, as context switches and interruptions can negatively impact productivity.
\end{itemize}

\subsection{LLMs and Predictive Analysis in Software Development}
LLMs are trained on vast amount of data that helps in various aspects by providing prediction by analysing historical data. The applications of LLMs for predicitive analysis in viewpoint of experts are given below:

\textit{Expert 1:} Code completion based on the overall code-base context can be accelerated rather than just syntax completion. Based on company-wide knowledge, OKRs can be set using these LLMs.

\subsection{Bias in LLMs for Software Development}
As LLMs are trained on a large amount of internet data, bias in its output is inevitable. This section discusses the biases present in different LLMs and how they can impact in the results.

\textit{Expert 5:} LLMs are generally trained on the large text data available on the internet. This data is largely human-generated or collected through systems created by humans \citep{ntoutsi2020bias}. Therefore, Artificial intelligence (AI) bias, also known as machine learning bias or algorithm bias, describes AI systems that generate biased outcomes that mirror and reinforce human prejudices in a community, including social inequity that exists today and has existed historically. The original training set, the algorithm, or the predictions the algorithm generates can all contain bias. It can be inherited in the form of gender stereotypes, religion, and racial prejudices \citep{kandpal2023large}. When producing text, the models can pick up on and replicate these biases that can result in discrimination in automated decision-making systems, among other negative outcomes, in addition to reinforcing negative perceptions \citep{gesi2023leveraging}. For example, Generative Pretrained Transformer (GPT)-3, which OpenAI released in 2020, showed an unparalleled capacity to produce coherent and contextually relevant text through training on a massive corpus of text data \citep{liyanage2023ethical}. Different studies reveal that content generated by GPT-3 is biased in many ways. A study investigates and reports brilliance bias in two models of GPT-3 \citep{shihadeh2022brilliance}. Also, another study findings show that some gender preconceptions in generated short stories that appear to be benign are reproduced by GPT-3 \citep{lorentzen2022social}. Similarly, the code generated by these LLMs can also produce unfair results, especially in handling bias sensitive tasks. Due to the increasing popularity of LLMs in software development, these biases may have far-reaching effects, including discriminatory hiring practices, lending choices in the financial sector and biased medical care. For example, when used in the recruitment process, the candidate's personal image and behavior are just as important as their level of professionalism. It is challenging for language models to accomplish this, whether it is over video chat or another form of communication. It requires specific analytical abilities, such as evaluating the tone of voice, gaze, and facial expressions of the candidate's response to ascertain whether the candidate is qualified for the position. Therefore, in order to minimize bias when using AI for recruiting, it also needs to be continuously monitored and the outcomes regularly analyzed by the relevant parties. In the event of issues, this could be more expensive than using conventional hiring techniques \citep{zhang2023impact}. Moreover, the preconceptions in the natural languages are reflected in the codes generated by these models if the task involves any protected attributes i.e., age, education, race, gender, city, region, and so on. For example, potential bias is detected in code snippets generated by five different state-of-the-art LLMs to assess employability given the attributes \citep{huang2023bias}. Also, studies investigating the replacement of human feedback with LLMs report verbosity bias i.e., the systems prefer long or more wordy answers than humans even if they appear of lower quality \citep{saito2023verbosity}. It can also lead to unnecessary complex and lengthy code generation for software development. Therefore, the software developed by these codes would also reflect bias in its results that can lead to unfair treatment to a special group, culture or region.

\textit{Expert 9:} LLMs have the power to dramatically influence and improve our lives. As the use of LLMs becomes more prevalent we must remain vigilant to ensure we properly assess all ethical considerations, particularly when these systems are applied within safety-critical domains, such as healthcare. There have been many reported incidents in recent years of LLMs producing biased or discriminatory results. Therefore, we must be extra vigilant in their use, particularly when they are deployed in situations where they could potentially cause harm. It is well documented that medicine has been developed for the white, male patient \citep{dresser1992wanted} and unfortunately, AI systems have the power to exacerbate this problem \citep{char2018implementing}. If we accept the world as it is, with all its flaws, AI systems such as LLMs will inevitably exhibit such behaviors and it is the most vulnerable patients in our society that will suffer the consequences. One of the fundamental, most popular uses of LLMs is as a chatbot – a mechanism for the user to engage with the model in a query-response interaction. The instant, and seemingly intelligent, responses an LLM can offer to individual queries has the potential to resemble a human conversation. While we know this is not true human interaction, it can appear genuine enough to simulate human communication. As such, these techniques have been proposed for use as healthcare chatbots, including some applications such as mental-health chatbots \citep{blease2023chatgpt}. There are clear potential benefits to the use of such technology for this type of application, for instance, a chatbot is always available, a chatbot does not get tired, a chatbot will never judge you. Users may find the non-judgmental, always-available support to be of great help, particularly if they feel stress at unsociable times such as during the night or in response to stressful situations where they do not have the opportunity to arrange to speak to a person. However, a chatbot is not a trained professional and there is a danger in allowing it too much influence over a person’s mood, particularly if that person is emotionally vulnerable. Any LLM or AI is trained on a dataset and they are trained according to their overall accuracy on results; such training does not consider the needs of an individual. A human counsellor or psychologist studies and trains for years to be able to detect and diagnose the ailment and correct treatment or response that a patient will need. How could we be sure an LLM will give the response needed by the individual and not merely offer a recommendation or platitude that it decides is statistically the best, based on a superficial attribute of the person such as their age, gender, or race? 
\paragraph  Mitigating AI biases is not a simple task, but that does not mean we should not aspire to address the issue and work towards a better future. The responsibility to ensure duty of care falls to each person who is involved in creating, deploying, and using a system that can be used to treat or influence a patient – including the developer, the product owners, and the institution or medical professional that employs any device that uses such technologies. While the use of AI techniques such as LLMs in medical and healthcare products is relatively new, there are a number of standards and guidelines that these stakeholders must familiarise themselves with. The European Union has recently reached a political agreement on the EU AI Act, to be finalized this year, which considers generative AI including LLMs. This was developed to ensure the regulated use of AI within the EU. NIST has also published a standard on identifying and managing bias in AI \citep{schwartztowards} and the IEEE plans to release the P7003 standard on Algorithmic Bias Considerations later this year (Koene et al., 2018) along with a certification course available through the IEEE CertifAIed \citep{certifAIEd}. Ideally, there would be a simple coding fix to this problem; we could merely install a plug-in, a function or an update that would ensure all software would act in a fair and non-discriminatory manner. Unfortunately, no such technological fix exists nor is it likely to ever exist. There are now a number of standards, guidelines, and regulations that developers must consider in order to ensure ethical and fair code generation, and it is up to each developer and user to make themselves aware of these guidelines and follow them to the best of their ability. If a LLM is employed as a chatbot that acts as a medical first-aider, or mental-health support bot, for instance, there is a duty to ensure it only offers fair, unbiased opinions or suggestions regardless of any sensitive or protected attribute from the user. LLMS are trained on real-world data and the real world is biased. This is a simple fact that we must not just accept, but must work to overcome. Technology is here and it is at our fingertips, but it is up to us to use it in a manner that helps all; we can only achieve this by considering the ethical and fairness implications of the use of these technologies, on all people, at each step of development and deployment.

\subsection{Ethical Considerations in LLMs for Software Development}
Other than bias, there are also some ethical concerns including privacy breaches associated with LLMs that need to be considered. The ethical implications of LLMs discussed by experts are as follows:

\textit{Expert 1:} At established scales such as enterprise customers, these models will have to deal with a lot of proprietary data bound by many data protection laws such as the General Data Protection Regulation (GDPR) in the European Union (EU). For best ethical practices, LLM service providers would need to comply with such laws e.g. through on-premises clusters/instances for those customers to ensure data locality and protection. Further, they would need to come to an agreement with customers about the kind of data that can be used to train their models both bespoke and general training.

\textit{Expert 5:} The extensive use of LLMs by students can limit their ability to learn, critical thinking, and experiment coding, and can raise ethical issues concerning cheating and plagiarism. The capacity of LLMs to enable plagiarism compromises academic integrity and undermines the goal of assessment, which is to impartially assess students' learning. Students who do excellent work using LLM models such as ChatGPT unfairly benefit from an advantage over their colleagues who do not have access to it. More importantly, using these models interferes with teachers' ability to effectively assess student performance, making it challenging to address students' learning issues \citep{cotton2023chatting}. One other aspect of biases in LLMs is the language dominancy that hinders the explanation of the problem for non-native speakers, thus producing faulty outputs. Moreover, it also imposes some other ethical concerns including breaching the privacy and confidentiality of individuals, dissemination of inaccurate information, and plagiarism. Italy was the first European country to ban ChatGPT by questioning the regulations for the storage and collection of users’ data. Authorities also expressed disapproval for the improper handling of erroneous information generated by the platform \citep{italyliftban}. Although the ban has been lifted now it raises some serious concerns about the LLMs model that is rigorously trained on text data without undergoing any pre-processing to preserve privacy, ensure transparency, mitigate biases, and handle inaccurate information that in code generation can limit assessability of functional correctness of the code generated \citep{liu2023your}. It becomes crucial to strike a balance between minimizing ethical hazards and utilizing LLMs' powers in sensitive contexts. This calls for strict regulations and a close examination of how they are used.

\subsection{Challenges of using LLMs in Software Development}
Although LLMs have huge potential but there are also some challenges that need to addressed as given below:

\textit{Expert 3:} There is a lot of hype around AI at the moment, largely due to the attention economy that is capitalizing on recent advancements like Palm 2, ChatGPT, Midjourney, and so on. Hype is not the same as recognizing the advancements in technology. Instead, the hype inflates expectations and confuses the real-world applications of AI with science fiction \citep{siegel2023ai}. This can impact a population of workers in a number of ways. It makes them emotional, scared, fatigued, and finally skeptical (when hyped-up predictions fail to come to pass). Many AI experts are skeptical of contemporary AI replacing software developers, highlighting how the most advanced AI systems today, LLMs, are just statistical models. These don’t have intelligence or understanding. Even if all software developers and companies used AI during development, the developer will still be required to design, review, and analyze the entire system. Rather than replacement, it is believed that it will be a tool to augment their work and abilities, which is known as Intelligence Augmentation \citep{planergy}. As the technology improves, more tasks and responsibilities of a software developer may be automated by AI. But, we don’t know if the capabilities of AI will be so advanced that they can replace humans entirely. a possible future where AI is advanced enough to replace a human at tasks will not be the same as the world is now. We can’t think about the future in terms of how the present is. In the same way that when the combustion engine was first invented, no one could have predicted how this would impact the world for the next century. While there were some paved roads, there was not an expanse of highways. Likewise, the secondary industries that exist due to cars – motorsports, petrol stations, mechanics, and aftermarkets for spare parts and accessories. No one could have predicted the vastness of the car industry at the time of its birth, and AI will be the same. There will be other opportunities that come with advanced AI. We have the benefit of learning from the past. Software developers should be aware that we are now in a revolutionary time, where the world is slowly starting to shift into a new trajectory.

\textit{Expert 7:} LLMs serve as a driving force behind the transformative change in developers' methodologies for conceptualizing, crafting, and enhancing code in software development endeavors. Similar to how automation rendered much of manual testing obsolete, the strategic utilization of LLMs holds promise for augmenting productivity and elevating the quality standards of software products. However, despite the proliferation of available options, many LLMs remain nascent in their capabilities, thus constraining their current applicability. Moreover, Instruction-tuned LLMs \citep{ouyang2022training} like OpenAI’s ChatGPT emerged as an early frontrunner in the conversational AI market, establishing widespread adoption until the introduction of Google’s Bard, a successor of Pathways Language Model 2 (PaLM 2) \citep{anil2023palm}. Utilizing both platforms extensively, developers tend to encounter a sense of disillusionment with ChatGPT's performance. Particularly within the context of interfacing with Windows services and applications, expectations were set for more dependable responses and proficient generation of code snippets tailored for Microsoft-owned libraries. While challenges without immediate resolutions are inherent in software development, the provision of inaccurate or irrelevant solutions is considered unacceptable. While prompting ChatGPT to suggest or create code snippets for web application development within the Microsoft Blazor framework \citep{blazor}, it was found the generated code block ostensibly addresses specific frontend properties that were absent in the framework. Furthermore, comparative evaluations between ChatGPT and Bard have highlighted instances where the latter has offered more appropriate solutions, intensifying dissatisfaction with the former's performance. 

\textit{Expert 8:} It is found that there is a lack of research on the
use of LLMs in software engineering, especially in areas like software testing,
turning natural language into code, and incorporating newer models like ChatGPT. Integrating LLMs into software engineering is complicated process. This
stresses the importance of conducting thorough literature reviews to understand
capabilities and limitations of LLMs which can bridge this knowledge gap \citep{hou2023large}. Despite the advantages, there are also several challenges and limitations when it comes to generating code using language models. The LLMs struggle with retaining the testing context which makes it difficult to generate accurate and relevant test scripts within the dynamic mobile app testing environment. LLMs may use inappropriate or deprecated APIs which can lead to non-executable scripts and necessitating manual intervention to align scripts with current APIs. It is also found that LLMs require human expertise for fine-tuning scripts to ensure compatibility with the mobile app’s environment, due to a lack of deep understanding of app intricacies.
LLMs have limited capabilities in generating scripts for complex test events like scrolling or managing pop-ups, requiring additional tools or manual efforts for comprehensive test coverage \citep{yu2023llm}. 

\textit{Expert 10:} LLM augmented development already looks set to accelerate rapidly in 2024, with new LLM models and tools being announced almost weekly.  Alongside the now well-established tools such as Copilot, Code Whisperer, and Gemini, other tools are regularly coming to market with new capabilities. However, some of the concerns about the proprietary models used to power these tools are they are all commercial, opaque, and require internet access.  Additionally, they still have unresolved questions about copyright.   A counter to these concerns is the growth of open-source models such as Llama, Mistral, and Dolphin which are opening up LLM development to far more participants in both industry and academia.  This will allow for far more research and innovation in the specialist area of AI coding tools.
While current coding assistants have proven very effective in code production, they are also still susceptible to excessive and irrelevant code as well as hallucinations.  They are also still limited in terms of software design \citep{liang2024large}.

\subsection{Future Directions and Innovations in LLMs for Software Development}
LLMs have also opened up various future opportunities for innovation to leverage the true potential of these architectures. Some of the opportunities from expert opinions are listed below:

\textit{Expert 1:} Currently, the biggest blocker is scaling these models. For an incremental improvement in the dataset size or efficiency, it requires a lot of computing power. This makes these extremely useful technologies a difficult sell at an enterprise level. Now that we’re approaching a saturation point in Moore’s law and the number of transistors we can fit in conventional silicon, we need to invest in silicon chips that follow the principles of a human brain. It will help democratize this breakthrough technology for the masses. The pricing models for such LLMs are based on the number of tokens used per prompt. As Prompt engineering is becoming an up-and-coming job profile, we can train engineers to be more efficient in their prompts and extract maximum values out of prompts. Currently, the feedback loop of these models for media-based inputs (such as pictures, videos, or audio) is long. With investment towards human-brain-like chips, we can reduce this feedback to reach near real-time response from LLMs for a wide array of input types.

\textit{Expert 6:} Despite the current productivity gains and performance of code language models, their full potential remains unrealised. Due to their unreliability, current models are primarily utilised for automating repetitive tasks and providing skeletal structures for ideas, simplifying the ideation process. However, a significant transformation in the software development pipeline is still on the horizon. Ethical considerations come into play as these code models could inadvertently aid in the development of malicious software, amplifying the ease of scaling such efforts. Future development efforts hinge on designing more reliable models. Key considerations include obtaining high-quality code data and fine-tuning language models to adhere to specific software development patterns and best practices, thereby minimising errors.

\textit{Expert 10:} The speed of evolution of new LLMs makes it difficult to predict exactly what technologies will be powering 3GL code generation tools, even in the near term. But what is easier to predict is the drivers of this technology, which remain relatively unchanged. Businesses will always aspire to convert business ideas to IT solutions with the lowest possible cost and time to market. The popularity of Low Code was on the back of this promise, but to date has not been able to deliver. AI tools may finally make it possible to convert business ideas to solutions with little or no need for traditional development. This could precipitate a realignment of solution delivery roles.   The role of developers who currently convert business requirements to code could move to no-code solutions, developed by ‘business engineers’ who are primarily aligned to business functions rather than IT.  These would be technical experts fully embedded in the business.  They may still require some traditional coding skills (for instance Excel has just embedded Python), but not in the way they do as application developers. Data scientists may increasingly occupy this space as many AI tools automating functions that data scientists currently need to code are coming to market. Operations has evolved significantly over the last decade with DevOps, but there is already signs this will evolve further and realign into Development and Platform Engineering,    Platform engineering is responsible for developing tooling to run infrastructure, production, and the development environment, abstracting away the complexity of operations from developers.  This is also a further enabler of moving solution development to business functions. At the same time, Platform engineers will require increasingly advanced AI coding tools. These will need all the development features of existing AI code, but in addition will need tools for IaC, configuration, security, testing, production management, and other pipeline toolsets. AI Ops skills may likely need this level of expertise to manage and deploy models. Finally, AI model developers and AI tool developers will require a range of existing and new AI tools to facilitate training, testing, and management of new models. Already these are using LLMs to generate new LLMs, and if this proves productive, no doubt new developer tools will emerge to accelerate these processes. It is possible the future of AI coding tools may be shaped as much by the future landscape of development roles in the AI age, as it is the evolution of the AI tools we have today.

\textit{Expert 11:}LLMs have numerous benefits, but they cannot completely replace human interaction. As a result, it is recommended that LLM models be used in conjunction with the manual code review process, with LLM handling mundane and repetitive tasks and human developers focusing on higher-order, creative problem-solving.

\section{Overview, Analysis and Discussion} \label{overview}

This section provides a brief overview of the major themes discussed by experts.

\subsection{Thematic Analysis}

After review of the expert opinions, 12 common themes have been identified to perform thematic analysis. They are as follows: Compliance, Ethics and Fairness, Lack of Innovation and Creativity, Human-AI Collaboration, Increase in Productivity, Code Review, Code Quality, NLG and NLU, Lack of Contextual Understanding, Collaboration and Inclusivity, Security Concerns, and Non-compliance to Industry Standards. Table \ref{tab:themes} displays the corresponding themes for each expert. 

The most common themes are `Ethics and Fairness' and `Code Quality' as out of 11 experts, 4 talk about the potential ethical considerations and fairness in application of LLMs in software development and their impact on code quality. Only one expert explicitly discusses the `NLG and NLU' capabilities of LLMs.  

\begin{table}[!h]
    \centering
      \small
        \caption{Major themes discussed by the Experts}
    \begin{tabular}{p{5.2cm}p{0.2cm}p{0.2cm}p{0.2cm}p{0.2cm}p{0.2cm}p{0.2cm}p{0.2cm}p{0.2cm}p{0.2cm}p{0.2cm}p{0.2cm}p{0.2cm}p{0.2cm}p{0.2cm}p{0.2cm}}
    \hline
       \textbf{Theme} & 1 &  2 & 3  & 4 & 5 & 6 & 7 & 8 & 9 & 10 & 11   \\
       \hline
       Compliance & & \checkmark & & & & &  \checkmark &\\
      
       Ethics and Fairness & \checkmark & & &  \checkmark  & \checkmark  & &  & & & \checkmark &\\
       Lack of innovation and creativity& & & & & \checkmark & & &  & &\\
       Human-AI collaboration& & & \checkmark & & & \checkmark& & & &\\
       Increase productivity & & & &  & & &  & & \checkmark& &  \\
       	Code review & \checkmark &  \checkmark &  & & & \checkmark& &   &\\
        Code quality & & & &   \checkmark &  & \checkmark& & \checkmark&\checkmark\\
       NLG and NLU &  \checkmark& & & & & & \\
       Lack of contextual understanding & & &   &  \checkmark  & & & & &\checkmark & \checkmark\\
       Collaboration and inclusivity	 & \checkmark & & & \checkmark & &   & &&&&  \checkmark\\
       Security concerns & \checkmark & & & & \checkmark & &\\
       Non-compliance & & & &  \checkmark && &  & &\checkmark\\
         \hline
         
    \end{tabular}
  
    \label{tab:themes}
\end{table}

\begin{table}[!ht]
    \centering
    \begin{tabular}{cc}
    \hline
        \textbf{Theme} & \textbf{Kappa score}  \\
        \hline
         
       Code Quality & 0.2142  \\
Code Review & 0.7441 \\
Collaboration and inclusivity & 0.7441 \\
Compliance &  1.0 \\
Ethics and fairness & 0.3773 \\
Human-AI collaboration & 0.6206 \\
Improve productivity & 0.2978\\
  Lack of contextual understanding & -0.1578\\
  Lack of innovation and creativity & 1.0\\
  NLG and NLU & -0.1379\\
  Non-compliance to industry standards & 1.0 \\
  Security concerns &  0.6206 \\
  \hline
  \textbf{Mean} & 0.5269 \\
       \hline
    \end{tabular}
    \caption{Cohen's Kappa For Inter-Annotator agreement between Themes}
    \label{tab:kappa}
\end{table}

Table \ref{tab:kappa} displays the kappa coefficients for the 12 thematic codes annotated by two annotators. The kappa score ranges from -1 to +1, with +1 conveying perfect agreement and -1 indicating less than chance agreement. `Compliance', ` Lack of innovation and creativity', and `Non-compliance to industry standards' have perfect agreement between the two annotators where as `Lack of contextual understanding' and `NLG and NLU' have less than chance agreement. The mean kappa score is 0.5269, which indicates moderate agreement between the annotator assigned labels.

\subsection{Opinions Overview}
\begin{figure}[!h]
\centering
  \includegraphics[scale=1]{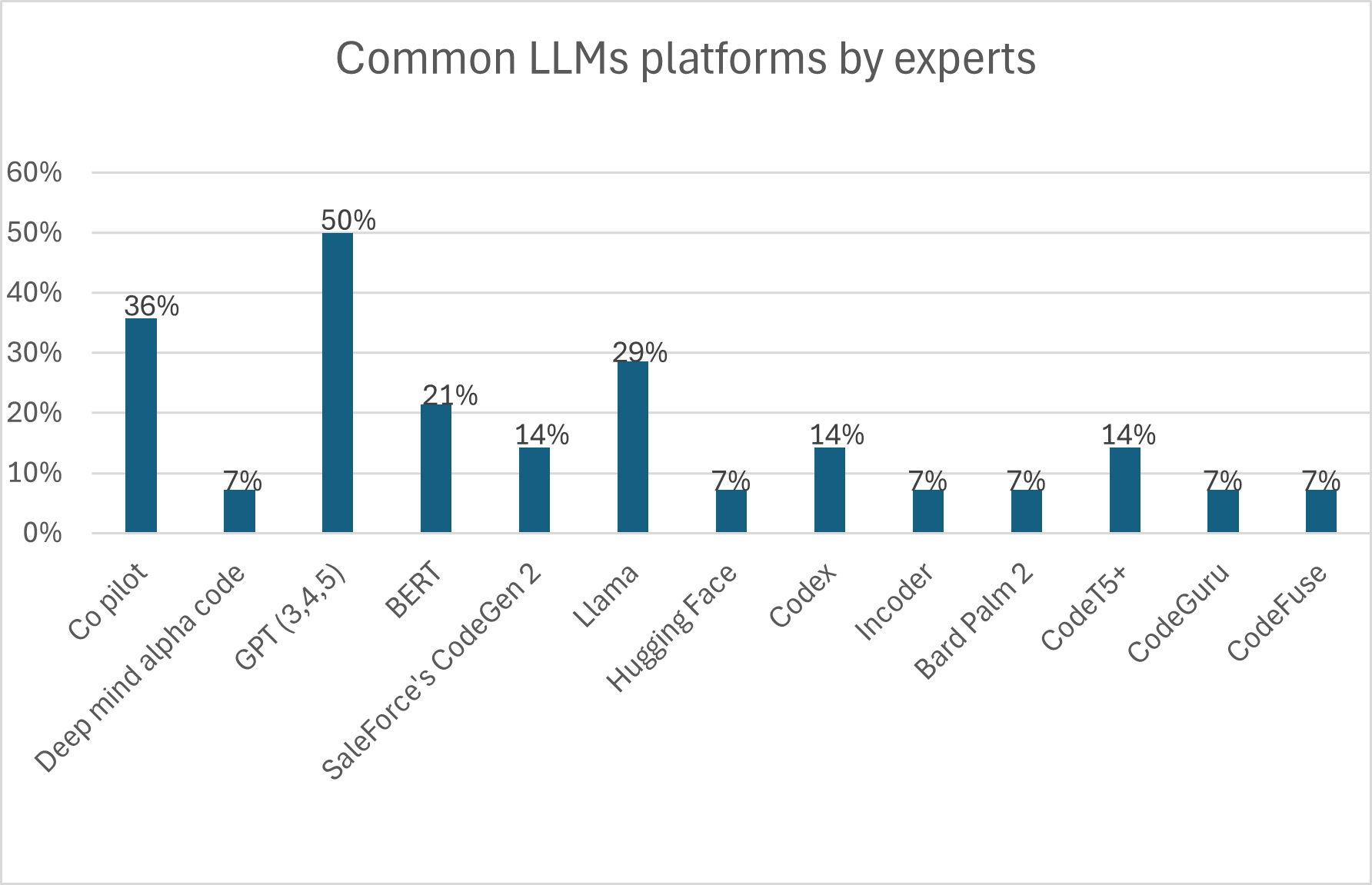}
  \caption{Most Common LLMs discussed by the Expert Panel}
  \label{fig:LLM}
\end{figure}

The role of LLMs in automating or assisting code generation is significant. Our expert panel has highlighted several advantages and use cases where LLMs can make a huge difference in software development. Four experts acknowledged that integrating these models in code generation activities can reduce the chances of errors in both syntax and logic, thus improving the code quality. Also, Experts 1 and 4 have mentioned that these platforms can aid cross-language collaboration i.e., code written in one programming language can be translated into other languages easily which increases usability and reduces effort. Moreover, Experts 4 and 8 mention that LLMs can generate code snippets based on statements written in plain languages or natural languages, increasing the LLMs' applicability span. In view of Expert 1, companies like Amazon Microsoft, Google, and Facebook (Meta) are big players in developing these platforms to enhance developer productivity and experience. Furthermore, Expert 8 draws attention to the ability of LLMs to create problem-specific software development or code generation that helps developers in efficient time utilization and reduces their burden. Fig.\ref{fig:LLM} shows the names of the most common LLMs discussed by the expert panel. GPT is the most frequent LLM mentioned, followed by Co pilot and Llama.

\paragraph  LLMs have greatly impacted the code review process. According to Expert 11  the incorporation of LLMs can enhance communication and collaboration between developers. An automated code review process can increase the effectiveness and accuracy of the code as well as the code review process. It can help to ensure best practices thus improving the overall code quality as recognized by Experts 1 and 2. Moreover, Expert 2, being a practitioner in the medical industry, says that the LLMs code review process can reduce the risk by mitigating defects in the early stage of medical device software development. Another important aspect that was brought forward by Expert 2 is related to security, as these models can also check codes for vulnerability detection and security audits. It can also help prioritize and optimize resource allocation.

\paragraph  Expert 1 also draws a view on the predictive analysis of LLMs. The data-driven predictions made by these language models can empower software development processes, help in optimized resource allocation, and provide assistance in long-term innovation and improvement in the software community.

\paragraph  One of the major concerns highlighted by experts are ethical and fairness issues as these models are trained on a large corpus of internet data. The data can be biased and these biases can be reflected in the form of gender stereotypes, religion, or social prejudice. The impact of these biases can highly affect a specific group when implemented in systems such as hiring software or medical care applications as mentioned by Experts 5 and 8. Other than that, the applications of open-source LLMs has raised ethical concerns for its extensive use by students. It limits the learning and innovation process. Furthermore, Experts 1 and 5 also highlight that these models can breach privacy and data protection laws as they can memorize new data. Experts 5 summarize that these models lack transparency and accountability. Also, the underlying bias can lead to discriminatory outputs.

\paragraph  There are several other limitations mentioned by the experts. These mainly include the inability of language models to handle ambiguity and vague instructions that make it difficult to generate correct, accurate, and relevant scripts. Also, LLMs are trained on the large text of the internet therefore, the code generated by them lacks specific industry standards or regulations. Also, these generated code snippets may not align with new APIs. Experts 3 and 11 express reservations about the applicability of these models, noting that viewing them as a replacement for human beings can bring anxiety and depression to the workforce. It is important to acknowledge that software development still requires humans to perform tasks such as higher-order planning, code verification, and fine-tuning of the code.

\paragraph  The full potential of LLMs is still not realized. Expert 1 says that these models should be trained on large datasets and also developers should be trained to use prompts efficiently to get the maximum from them. Expert 6 believe that these models should be trained to increase the reliability of the content generated by them by increasing data exposure and testing protocols.

\subsection{Discussion} \label{discussion}
In this section, we will discuss the findings of the study with respect to the research questions. 

\begin{enumerate}
    \item RQ1: What are the applications and advantages of LLMs in software development as per expert opinions?

    The thematic analysis approach applied in this study has revealed several perceived benefits that the integration of LLMs in software development can provide. LLMs have been used to perform several tasks in software development, such as code reviewing, pair programming \citep{wu2023ai}, code generation, converting from natural language prompt to code, generating documentation for the code, improving code quality \citep{hou2023large}, translating code across programming languages, and predicting code-snippets in real time. LLM-assisted platforms such as GitHub Copilot can serve as valuable assistants for developers by automating tedious tasks and allowing them to focus on tasks that require creative problem solving \citep{githubcopilot}. Within a pre-defined framework, LLMs can help ensure compliance to standards \citep{hou2023large}. They can also help in maintaining version control for the various dependencies in an application. LLMs can also support cross-platform and cross-lingual collaboration between developers.

    \item RQ2: What are the limitations and potential concerns of LLMs in software development as per expert opinions? 
    
    Despite seeing extensive application in software development, there are still significant drawbacks to using LLMs as affirmed by the expert opinions. Firstly, LLMs suffer from `hallucinations', which refers to generating factually incorrect information \citep{yao2023llm}. Therefore, responses provided by LLMs should always be verified. Secondly, over-reliance on LLMs can limit the critical thinking skills in new developers and hamper innovation. Students can also unfairly use LLMs to complete assignments or tests \cite{perkins2023academic}. Vague instructions given to LLMs may result in incorrect scripts. There are also significant ethical considerations to the application of LLMs in software development \citep{liyanage2023ethical}. Deep learning algorithms contain inherent biases which may get amplified with downstream applications. LLMs are pre-trained on large amount of data, which also contains biases such as gender bias \citep{dong2024disclosure}, racial bias, and brilliance bias. Integration of LLMs into software development workflows may also disrupt version control and compliance, model transparency, and responsible AI practices. Additional training may also be required to efficiently use the LLMs.
    
\end{enumerate}

Based on the analysis of the expert opinions performed in the study, LLMs have the ability to efficiently perform tasks in the software development lifecycle. However, their integration into existing workflows must be done while keeping the various ethical, security, and privacy concerns in mind. The study analysed opinions belonging to experts from a varied demographic, and from the industry and academia sectors. However, additional work must be conducted by considering different populations to ensure that the findings of the study generalise to other groups. Exploring additional tools such as in-person interviews and structured surveys, may allow an in-depth analysis of the opinions of experts in the field of software development and NLP. 

\section{Conclusion and Future Work}  \label{conclusion}

Since the launch of LLMs, they have been utilised for various applications such as image generation, healthcare, education, and software development. LLMs such as CodeBERT and Codex can be used for various tasks in the software development lifecycle such as code review, generation of documentation, pair programming, and so on. Despite the numerous advantages, there are several limitations as well as ethical implications to their use, and any integration with existing frameworks needs to be carefully monitored. The thematic analysis of the expert opinions performed in this study reveals the advantages and limitations of using LLMs for software development. Moving forward, it is important to understand that LLMs are based on algorithms that contain bias, which can be further propagated through irresponsible use. The future scope of this work includes deeper thematic analysis through approaches such as keyword and sentiment analysis.

\section*{Acknowledgement}
The project is partially supported by the DkIT Postgraduate Scholarship, Taighde Éireann – Research Ireland under Grant number 13/RC/2094\_2, and Grant number 21/FFP-A/9255.


\bibliography{references}
\bibliographystyle{plainnat}

\end{document}